\documentclass[10pt,twocolumn,letterpaper]{article}

\usepackage{iccv}
\usepackage{times}
\usepackage{epsfig}
\usepackage{graphicx}
\usepackage{amsmath}
\usepackage{amssymb}
\usepackage{float}
\usepackage{enumitem}

\setlist[enumerate]{leftmargin=*, itemsep=-1mm}
\setlist[itemize]{leftmargin=*, itemsep=-1mm}

\usepackage{subcaption}

\iccvfinalcopy 

\def\iccvPaperID{603} 

\ificcvfinal\fi
\begin{document}


\title{Large Pose 3D Face Reconstruction from a Single
  Image via Direct Volumetric CNN Regression}

\author{Aaron S. Jackson$^1$\hspace{1cm}
  Adrian Bulat$^1$\hspace{1cm}
  Vasileios  Argyriou$^2$\hspace{1cm}
  Georgios Tzimiropoulos$^1$\hspace{1cm} \\
  \\
  $^1$ The University of Nottingham, UK\hspace{1cm}
  $^2$ Kingston University, UK \\
  \\
  $^1$\{aaron.jackson, adrian.bulat, yorgos.tzimiropoulos\}@nottingham.ac.uk \\
  $^2$ vasileios.argyriou@kingston.ac.uk
}
\twocolumn[{%
\renewcommand\twocolumn[1][]{#1}%
\maketitle
\begin{center}
    \centering
    \includegraphics[width=\linewidth]{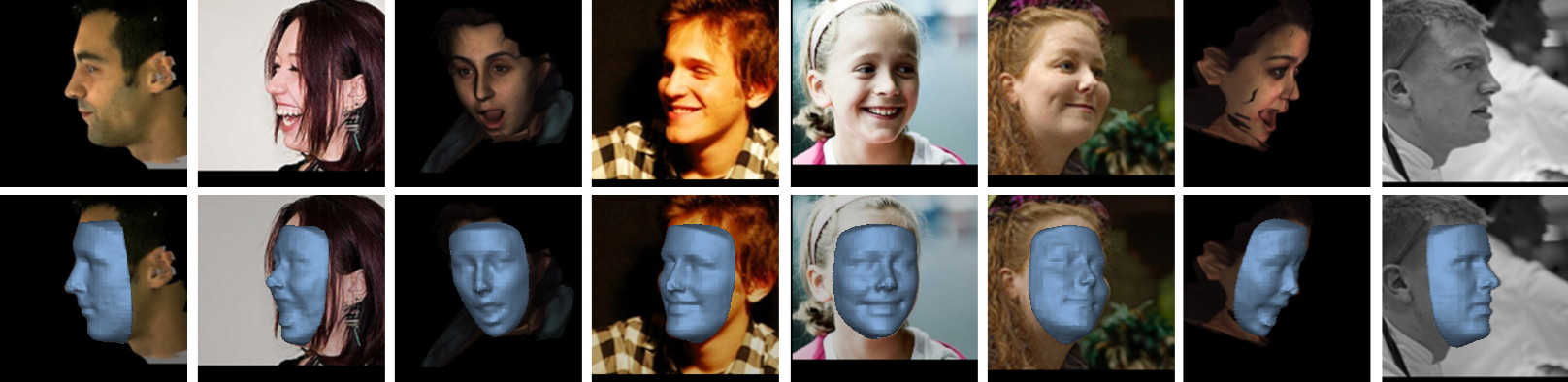}
    \captionof{figure}{A few results from our \textit{VRN - Guided} method,
    on a full range of pose, including large expressions.}
  \label{fig:preview}
\end{center}%
}]
\maketitle

\begin{abstract}
  3D face reconstruction is a fundamental Computer Vision problem of
  extraordinary difficulty. Current systems often assume the
  availability of multiple facial images (sometimes from the same
  subject) as input, and must address a number of methodological
  challenges such as establishing dense correspondences across large
  facial poses, expressions, and non-uniform illumination. In general
  these methods require complex and inefficient pipelines for model
  building and fitting. In this work, we propose to address many of
  these limitations by training a Convolutional Neural Network (CNN)
  on an appropriate dataset consisting of 2D images and 3D facial
  models or scans. Our CNN works with just a single 2D facial image,
  does not require accurate alignment nor establishes dense
  correspondence between images, works for arbitrary facial poses and
  expressions, and can be used to reconstruct the whole 3D facial
  geometry (including the non-visible parts of the face) bypassing the
  construction (during training) and fitting (during testing) of a 3D
  Morphable Model. We achieve this via a simple CNN architecture that
  performs direct regression of a volumetric representation of the 3D
  facial geometry from a single 2D image. We also demonstrate how the
  related task of facial landmark localization can be incorporated
  into the proposed framework and help improve reconstruction quality,
  especially for the cases of large poses and facial
  expressions. Code and models will be made available at \verb|http://aaronsplace.co.uk|
\end{abstract}
\vspace{-4mm}

\section{Introduction}
3D face reconstruction is the problem of recovering the 3D facial
geometry from 2D images. Despite many years of research, it is still
an open problem in Vision and Graphics research. Depending on the
setting and the assumptions made, there are many variations of it as
well as a multitude of approaches to solve it. This work is on 3D face
reconstruction using only a single image. Under this setting, the
problem is considered far from being solved. In this paper, we propose
to approach it, for the first time to the best of our knowledge, by
directly learning a mapping from pixels to 3D coordinates using a
Convolutional Neural Network (CNN). Besides its simplicity, our
approach works with totally unconstrained images downloaded from the
web, including facial images of arbitrary poses, facial expressions
and occlusions, as shown in Fig.~\ref{fig:preview}. \newline
\textbf{Motivation.} No matter what the underlying assumptions are,
what the input(s) and output(s) to the algorithm are, 3D face
reconstruction requires in general complex pipelines and solving
non-convex difficult optimization problems for both model building
(during training) and model fitting (during testing). In the following
paragraph, we provide examples from 5 predominant approaches:

\begin{enumerate}
\item In the 3D Morphable Model (3DMM) \cite{blanz1999morphable,
    romdhani2005estimating}, the most popular approach for estimating
  the full 3D facial structure from a single image (among others),
  training includes an iterative flow procedure for dense image
  correspondence which is prone to failure. Additionally, testing requires a
  careful initialisation for solving a difficult highly non-convex
  optimization problem, which is slow.
\item
The work of \cite{kemelmacher20113d}, a popular approach for 2.5D reconstruction from a single image, formulates and solves a carefully initialised (for frontal images only) non-convex optimization problem for recovering the lighting, depth, and albedo in an alternating manner where each of the sub-problems is a difficult optimization problem per se.
\item
In \cite{kemelmacher2011face}, a quite popular recent approach for creating a neutral subject-specific 2.5D model from a near frontal image, an iterative procedure is proposed which entails localising facial landmarks, face frontalization, solving a photometric stereo problem, local surface normal estimation, and finally shape integration.
\item
In \cite{suwajanakorn2014total}, a state-of-the-art pipeline for reconstructing a highly detailed 2.5D facial shape for each video frame, an average shape and an illumination subspace for the specific person is firstly computed (offline), while testing is an iterative process requiring a sophisticated pose estimation algorithm, 3D flow computation between the model and the video frame, and finally shape refinement by solving a shape-from-shading optimization problem.
\item
More recently, the state-of-the-art method of \cite{roth2016adaptive} that produces the average (neutral) 3D face from a collection of personal photos, firstly performs landmark detection, then fits a 3DMM using a sparse set of points, then solves an optimization problem similar to the one in \cite{kemelmacher2011face}, then performs surface normal estimation as in \cite{kemelmacher2011face} and finally performs surface reconstruction by solving another energy minimisation problem.
\end{enumerate}

Simplifying the technical challenges involved in the
aforementioned works is the main motivation of this paper.

\subsection{Main contributions}
We describe a very simple approach which bypasses many of the
difficulties encountered in 3D face reconstruction by using a
novel volumetric representation of the 3D facial geometry, and
an appropriate CNN architecture that is trained to regress directly
from a 2D facial image to the corresponding 3D volume. An overview of
our method is shown in Fig.~\ref{fig:cnnall}. In summary, our contributions
are:
\begin{itemize}
\item Given a dataset consisting of 2D images and 3D face scans, we
  investigate whether a CNN can learn directly, in an end-to-end fashion, the
  mapping from image pixels to the full 3D facial structure geometry
  (including the non-visible facial parts). Indeed, we show that the
  answer to this question is positive.
\item We demonstrate that our CNN works with just a single 2D facial
  image, does not require accurate alignment nor establishes dense
  correspondence between images, works for arbitrary facial poses and
  expressions, and can be used to reconstruct the whole 3D facial
  geometry bypassing the construction (during training) and fitting
  (during testing) of a 3DMM.
\item We achieve this via a simple CNN architecture that performs
  \textit{direct} regression of a volumetric representation of the 3D
  facial geometry from a single 2D image. 3DMM fitting is not
  used. Our method uses only 2D images as input to the proposed CNN
  architecture.
\item We show how the related task of 3D facial landmark localisation
  can be incorporated into the proposed framework and help improve
  reconstruction quality, especially for the cases of large poses and
  facial expressions.
\item We report results for a large number of experiments on both
  controlled and completely unconstrained images from the web,
  illustrating that our method outperforms prior work on single image
  3D face reconstruction by a large margin.
\end{itemize}


\section{Closely related work}

This section reviews closely related work in 3D face reconstruction,
depth estimation using CNNs and work on 3D representation modelling with CNNs.

\textbf{3D face reconstruction.} A full literature review of 3D face
reconstruction falls beyond the scope of the paper; we simply note
that our method makes minimal assumptions i.e. it requires just a
single 2D image to reconstruct the full 3D facial structure, and works
under arbitrary poses and expressions. Under the single image setting,
the most related works to our method are based on 3DMM fitting
\cite{blanz1999morphable, romdhani2005estimating, zhu2016face,
  jourabloo2016large, huber2016multiresolution} and the work of
\cite{liu2016joint} which performs joint face reconstruction and
alignment, reconstructing however a neutral frontal face.

The work of \cite{romdhani2005estimating} describes a multi-feature
based approach to 3DMM fitting using non-linear least-squares
optimization (Levenberg-Marquardt), which given appropriate
initialisation produces results of good accuracy. More recent work has
proposed to estimate the update for the 3DMM parameters using CNN
regression, as opposed to non-linear optimization.  In
\cite{jourabloo2016large}, the 3DMM parameters are estimated in six
steps each of which employs a different CNN. Notably,
\cite{jourabloo2016large} estimates the 3DMM parameters on a sparse
set of landmarks, i.e. the purpose of \cite{jourabloo2016large} is 3D
face alignment rather than face reconstruction. The method of
\cite{zhu2016face} is currently considered the state-of-the-art in
3DMM fitting. It is based on a single CNN that is iteratively applied to
estimate the model parameters using as input the 2D image and a
3D-based representation produced at the previous iteration.
Finally, a state-of-the-art cascaded regression landmark-based
3DMM fitting method is proposed in~\cite{huber2016multiresolution}.

Our method is different from the aforementioned methods in the following ways:
\begin{itemize}
\item Our method is direct. It does not estimate 3DMM parameters and,
  in fact, it completely bypasses the fitting of a 3DMM. Instead, our
  method directly produces a 3D volumetric representation of the
  facial geometry.
\item
Because of this fundamental difference, our method is also radically different in terms of the CNN architecture used: we used one that is able to make spatial predictions at a voxel level, as opposed to the networks of \cite{zhu2016face, jourabloo2016large} which holistically predict the 3DMM parameters.
\item
Our method is capable of producing reconstruction results for completely unconstrained facial images from the web covering the full spectrum of facial poses with arbitrary facial expression and occlusions. When compared to the state-of-the-art CNN method for 3DMM fitting of \cite{zhu2016face}, we report large performance improvement.
\end{itemize}


Compared to works based on shape from shading \cite{kemelmacher20113d, suwajanakorn2014total}, our method cannot capture such fine details. However, we believe that this is primarily a problem related to the dataset used rather than of the method. Given training data like the one produced by \cite{kemelmacher20113d, suwajanakorn2014total}, then we believe that our method has the capacity to learn finer facial details, too.

\textbf{CNN-based depth estimation.} Our work has been inspired by the
work of \cite{eigen2015predicting, eigen2014depth} who showed that a
CNN can be directly trained to regress from pixels to depth values
using as input a single image. Our work is different from
\cite{eigen2015predicting, eigen2014depth} in 3 important respects:
Firstly, we focus on faces (i.e. deformable objects) whereas
\cite{eigen2015predicting, eigen2014depth} on general scenes
containing mainly rigid objects. Secondly, \cite{eigen2015predicting,
  eigen2014depth} learn a mapping from 2D images to 2D depth maps,
whereas we demonstrate that one can actually learn a mapping from 2D
to the full 3D facial structure including the non-visible part of the
face. Thirdly, \cite{eigen2015predicting, eigen2014depth} use a
multi-scale approach by processing images from low to high
resolution. In contrast, we process faces at fixed scale (assuming
that this is provided by a face detector), but we build our CNN based
on a state-of-the-art bottom-up top-down module
\cite{newell2016stacked} that allows analysing and combining CNN
features at different resolutions for eventually making predictions at
voxel level.

\textbf{Recent work on 3D.} We are aware of only one
work which regresses a volume using a CNN. The work
of~\cite{choy20163d} uses an LSTM to regress the 3D structure of
multiple object classes from one or more images. This is different from our work in at least two ways. Firstly, we treat our reconstruction as a semantic segmentation problem by regressing a volume which is spatially aligned with the image. Secondly, we work from only one image in one single step, regressing a much larger volume of $192\times 192\times 200$ as opposed to the $32\times 32\times 32$ used in~\cite{choy20163d}. The work of \cite{tulsiani2016learning} decomposes an input 3D shape into shape primitives which along with a set of  parameters can be used to re-assemble the given shape. Given the input shape, the goal of \cite{tulsiani2016learning} is to regress the shape primitive parameters which is achieved via a CNN. The method of \cite{pavlakos2016coarse} extends classical work on heatmap regression \cite{tompson2015efficient, pfister2015flowing} by proposing a 4D representation for regressing the location of \textit{sparse} 3D landmarks for human pose estimation. Different from \cite{pavlakos2016coarse}, we demonstrate that a 3D volumetric representation is particular effective for learning \textbf{dense} 3D facial geometry. In terms of 3DMM fitting, very recent work includes  \cite{richardson2016learning} which uses a CNN similar to the one of \cite{zhu2016face} for producing coarse facial geometry but additionally includes a second network for refining the facial geometry and a novel rendering layer for connecting the two networks. Another recent work is \cite{tran2016regressing} which uses a very deep CNN for 3DMM fitting.


\section{Method}


This section describes our framework including the proposed data representation used.

\subsection{Dataset}

Our aim is to regress the full 3D facial structure from a 2D image. To
this end, our method requires an appropriate dataset consisting of 2D
images and 3D facial scans. As our target is to apply the method on
completely unconstrained images from the web, we chose the dataset of
\cite{zhu2016face} for forming our training and test sets. The dataset
has been produced by fitting a 3DMM built from the combination of the
Basel \cite{paysan20093d} and FaceWarehouse
\cite{cao2014facewarehouse} models to the unconstrained images of the
300W dataset \cite{sagonas2013semi} using the multi-feature fitting
approach of \cite{romdhani2005estimating}, careful initialisation and
by constraining the solution using a sparse set of landmarks. Face
profiling is then used to render each image to 10-15 different poses
resulting in a large scale dataset (more than 60,000 2D facial images
and 3D meshes) called 300W-LP. Note that because each mesh is
produced by a 3DMM, the vertices of all produced meshes are in dense
correspondence; however this is not a prerequisite for our method and
unregistered raw facial scans could be also used if available
(e.g. the BU-4DFE dataset~\cite{yin2008high}).

\subsection{Proposed volumetric representation}

Our goal is to predict the coordinates of the 3D vertices of each
facial scan from the corresponding 2D image via CNN regression. As a
number of works have pointed out (see for example
\cite{tompson2015efficient, pfister2015flowing}), direct regression of
all 3D points concatenated as a vector using the standard L2 loss
might cause difficulties in learning because a single correct value
for each 3D vertex must be predicted. Additionally, such an approach
requires interpolating all scans to a vector of a fixed dimension, a
pre-processing step not required by our method. Note that similar
learning problems are encountered when a CNN is used to regress model
parameters like the 3DMM parameters rather than the actual
vertices. In this case, special care must be taken to weight parameters
appropriately using the Mahalanobis distance or in general some
normalisation method, see for example \cite{zhu2016face}. We compare
the performance of our method with that of a similar method~\cite{zhu2016face} in Section~\ref{S:Results}.

\begin{figure}
  \centering
  \includegraphics[width=\linewidth]{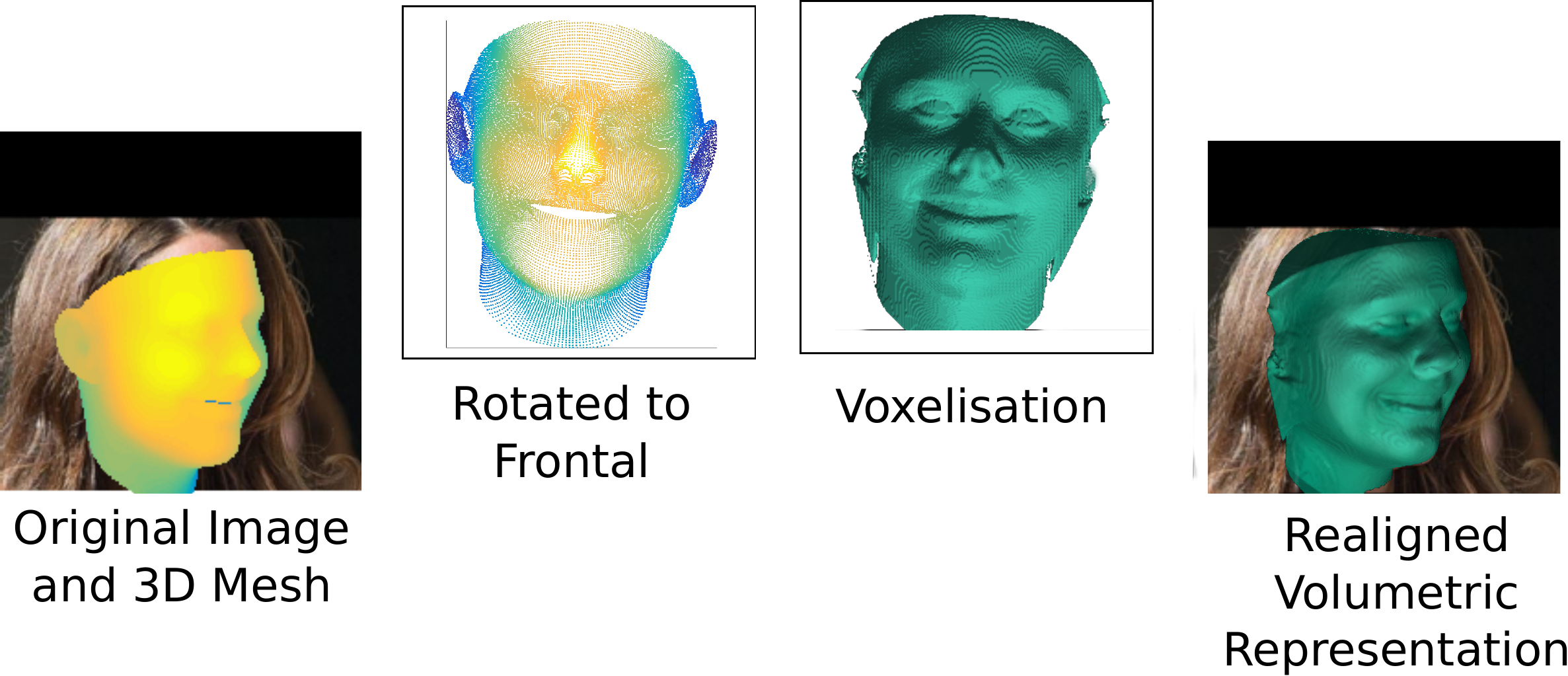}
  \caption{The voxelisation process creates a volumetric representation of the 3D face mesh, aligned with the 2D image.}
  \label{fig:discretisation}
\end{figure}

\begin{figure}
\includegraphics[width=\linewidth]{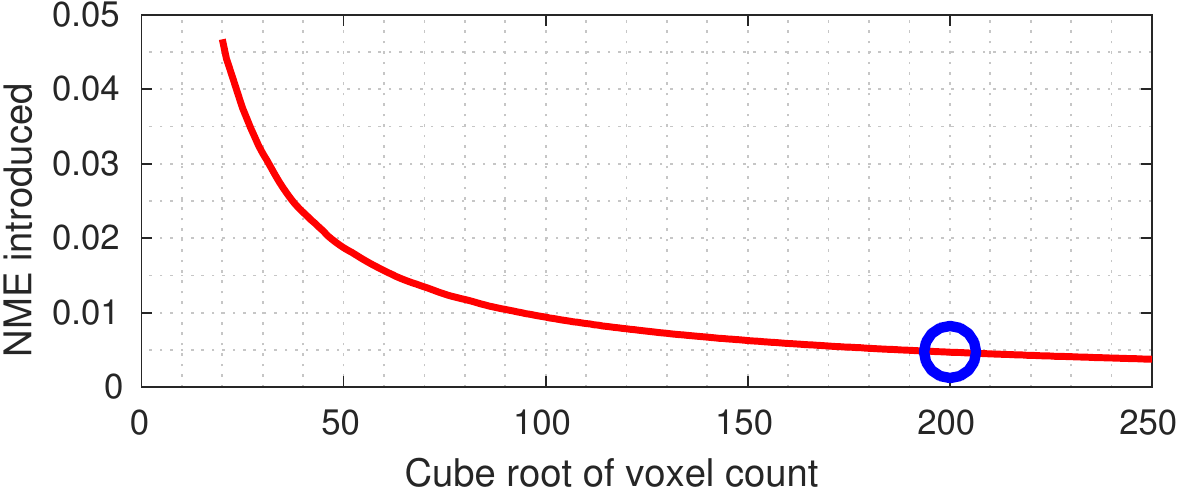}
\caption{The error introduced due to voxelisation, shown as a function of volume density.}
\label{fig:voxerror}
\end{figure}

To alleviate the aforementioned learning problem, we propose to
reformulate the problem of 3D face reconstruction as one of 2D to 3D
image segmentation: in particular, we convert each 3D facial scan into 
a 3D binary volume $\mathbf{V}_{whd}$ by discretizing the 3D space
into voxels $\{w,h,d\}$, assigning a value of 1 to all
points enclosed by the 3D facial scan, and 0 otherwise. That is to say
$ V_{whd}$ is the ground truth for voxel $\{w,h,d\}$ and is equal to
1, if voxel $\{w,h,d\}$ belongs to the 3D volumetric representation of
the face and 0 otherwise (i.e. it belongs to the background). The
conversion is shown in Fig.~\ref{fig:discretisation}. Notice that the process creates a volume fully aligned with the 2D image. The importance of spatial alignment is analysed in more detail in Section~\ref{sec:spatialimportance}. The error caused by
discretization for a randomly picked facial scan as a function of the
volume size is shown in Fig.~\ref{fig:voxerror}. Given that the error of
state-of-the-art methods
\cite{roth2016adaptive,liu2016joint} is of the order of a few mms, we
conclude that discretization by $192\times 192\times 200$ produces negligible
error.

Given our volumetric facial representation, the problem of regressing
the 3D coordinates of all vertices of a facial scan is reduced to one
of 3D binary volume segmentation. We approach this problem using
recent CNN architectures from semantic image segmentation
\cite{long2015fully} and their extensions \cite{newell2016stacked}, as
described in the next subsection.

\subsection{Volumetric Regression Networks}

\begin{figure*}
  \centering
  \begin{subfigure}[t]{1\textwidth}
  \centering
  \includegraphics[width=0.81\linewidth]{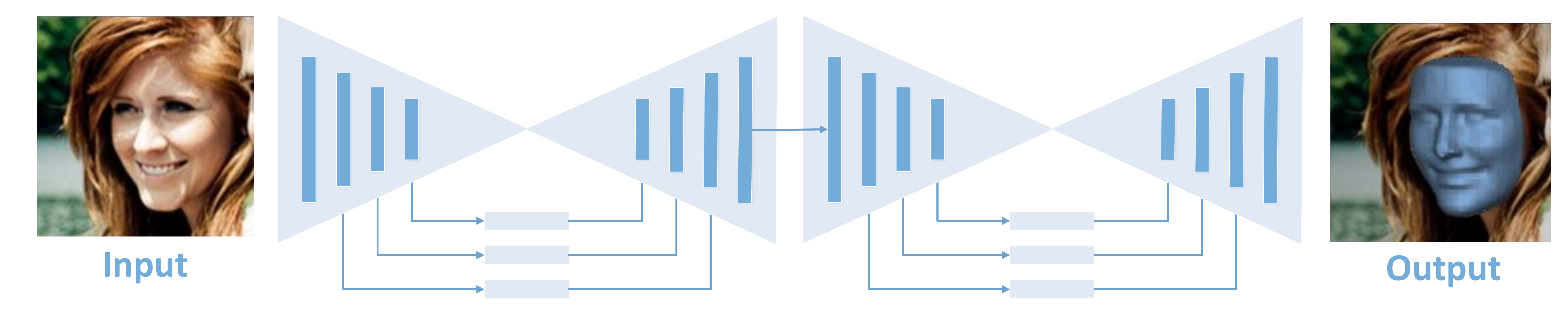}
  \caption{The proposed \textit{Volumetric Regression Network (VRN)} accepts as input an RGB input and directly regresses a 3D volume completely bypassing the fitting of a 3DMM. Each rectangle is a residual module of 256 features.}

  \label{fig:cnnbaseline}
  \end{subfigure}
  ~
   \begin{subfigure}[t]{0.98\textwidth}
     \centering
  \includegraphics[width=\linewidth]{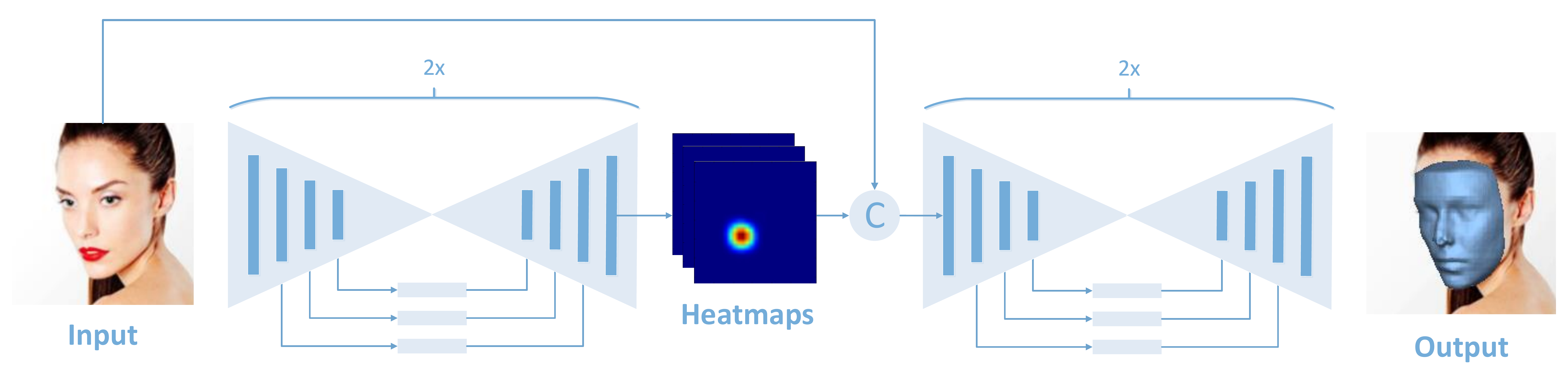}
  \caption{The proposed \textit{VRN - Guided} architecture firsts detects the 2D projection of the 3D landmarks, and stacks these with the original image. This stack is fed into the reconstruction network, which directly regresses the volume.}
  \label{fig:guidednet}
   \end{subfigure}
   ~
   \begin{subfigure}[t]{1\textwidth}
   \centering
  \includegraphics[width=0.6\linewidth]{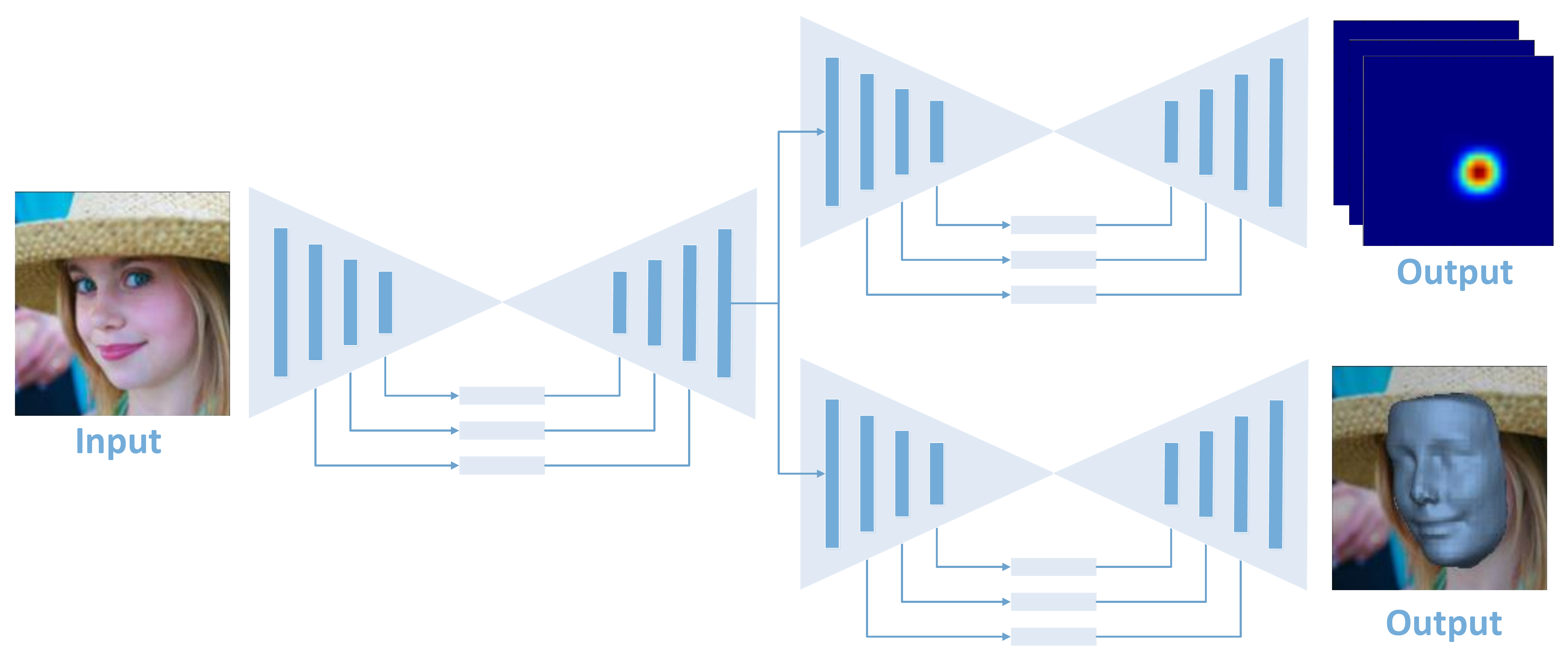}
  \caption{The proposed \textit{VRN - Multitask} architecture regresses both the 3D facial volume and a set of sparse facial landmarks.}
  \label{fig:cnnmultitask}
  \vspace{-2mm}
  \end{subfigure}
  \caption{An overview of the proposed three architectures for Volumetric Regression: \textit{Volumetric Regression Network (VRN)}, \textit{VRN - Guided} and \textit{VRN - Multitask}.}
  \label{fig:cnnall}
  \vspace{-5mm}
\end{figure*}

In this section, we describe the proposed volumetric regression network, exploring several architectural variations described in detail in the following subsections:

\textbf{Volumetric Regression Network (VRN)}. We wish to learn a mapping from the 2D facial image to its
corresponding 3D volume $f: \mathbf{I} \rightarrow \mathbf{V}$. Given
the training set of 2D images and constructed volumes, we learn this mapping using a CNN. Our CNN
architecture for 3D segmentation is based on the ``hourglass
network'' of \cite{newell2016stacked} an extension of the fully
convolutional network of \cite{long2015fully} using skip connections
and residual learning \cite{he2015deep}. Our volumetric architecture consists of two hourglass
modules which are stacked together without intermediate
supervision. The input is an RGB image and the output is a volume of $192\times 192\times 200$ of real values. This architecture is
shown in Fig.~\ref{fig:cnnbaseline}. As it can be observed, the
network has an encoding/decoding structure where a set of convolutional layers are firstly used to compute a feature representation of fixed dimension. This representation is further processed back to the spatial domain, re-establishing spatial correspondence between the input image and the output volume. Features are hierarchically
combined from different resolutions to make per-pixel predictions. The second hourglass is used to refine this output, and has an identical structure to that of the first one.

We train our volumetric regression network using the sigmoid cross entropy loss function:
\begin{equation}
  l_{1} = \sum\limits_{w=1}^{W} \sum\limits_{h=1}^{H}\sum\limits_{d=1}^{D}[V_{whd}\log \widehat{V}_{whd}+(1-V_{whd})\log(1-\widehat{V}_{whd})],
\end{equation}
where $\widehat{V}_{whd}$ is the corresponding sigmoid output at voxel $\{w,h,d\}$ of the regressed volume.

\begin{figure}
  \centering
  \includegraphics[width=0.4\linewidth]{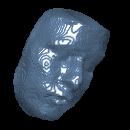}
  \includegraphics[width=0.4\linewidth]{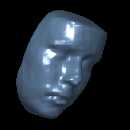}
  \caption{Comparison between making hard (binary) vs soft (real) predictions. The latter produces a smoother result.}
  \label{fig:roughvssmooth}
  \vspace{-4mm}
\end{figure}

At test time, and given an input 2D image, the network regresses a 3D volume from which the outer 3D facial mesh is recovered. Rather than making hard (binary) predictions at pixel level, we found that the soft sigmoid output is more useful for further processing. Both
representations are shown in Fig.~\ref{fig:roughvssmooth} where
clearly the latter results in smoother results. Finally, from the 3D volume, a mesh can be formed by generating the iso-surface of the volume. If needed, correspondence between this variable length mesh and a fixed mesh can be found using Iterative Closest Point (ICP).

\textbf{VRN - Multitask}. We also propose a Multitask VRN, shown in Fig.~\ref{fig:cnnmultitask}, consisting of three hourglass modules. The first hourglass provides features to a fork of two hourglasses. The first of this fork regresses the 68 iBUG landmarks \cite{sagonas2013semi} as 2D Gaussians, each on a separate channel. The second hourglass of this fork directly regresses the 3D structure of the face as a volume, as in the aforementioned
unguided volumetric regression method. The goal of this multitask network is to learn more reliable features which are better suited to the two tasks.

\textbf{VRN - Guided}. We argue that reconstruction should benefit from firstly performing a simpler face analysis task; in particular we propose an architecture for
volumetric regression guided by facial landmarks. To this end, we train a stacked hourglass network which
accepts guidance from landmarks during training and
inference. This network has a similar architecture to the unguided volumetric regression method, however the input to this architecture is an RGB image stacked with 68 channels, each containing a Gaussian ($\sigma = 1$, approximate diameter of 6 pixels) centred on each of
the 68 landmarks. This stacked representation and architecture is demonstrated in Fig.~\ref{fig:guidednet}. During training we used the ground truth landmarks while during testing we used a stacked hourglass network trained for facial landmark localisation. We call this network \textit{VRN - Guided}.

\subsection{Training}

Each of our architectures was trained end-to-end using RMSProp with an initial learning rate of $10^{-4}$, which was lowered after 40 epochs to $10^{-5}$. During training, random augmentation was applied to each
input sample (face image) and its corresponding target (3D volume): we applied in-plane rotation $r\in[-45^{\circ}, ..., 45^{\circ}]$, translation  $t_z,t_y\in[-15,...,15]$ and scale
$s\in [0.85,...,1.15]$ jitter. In 20\% of cases, the input and target were flipped horizontally. Finally, the input samples were adjusted with some colour scaling on each RGB channel. 

In the case of the \textit{VRN - Guided}, the landmark detection module was trained to regress Gaussians with standard deviation of approximately 3 pixels ($\sigma = 1$). 



\section{Results} \label{S:Results}

\begin{figure}
\centering
\includegraphics[width=\linewidth]{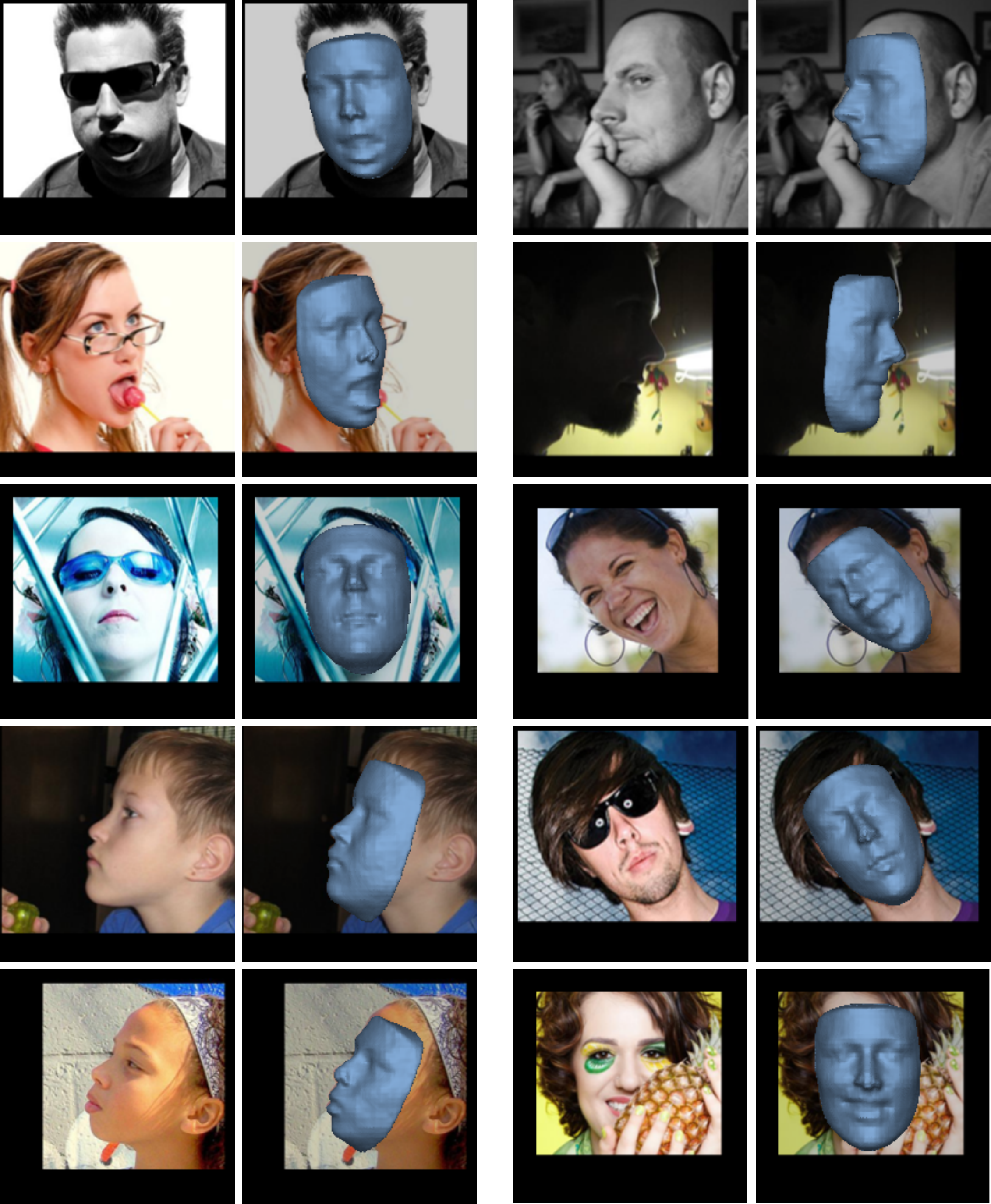}
\caption{Some visual results from the AFLW2000-3D dataset generated using our \textit{VRN - Guided} method.}
\label{fig:aflw2000res}
\vspace{-4mm}
\end{figure}

\begin{figure*}
\centering
\begin{subfigure}[t]{0.43\textwidth}
\centering
  \includegraphics[width=\linewidth]{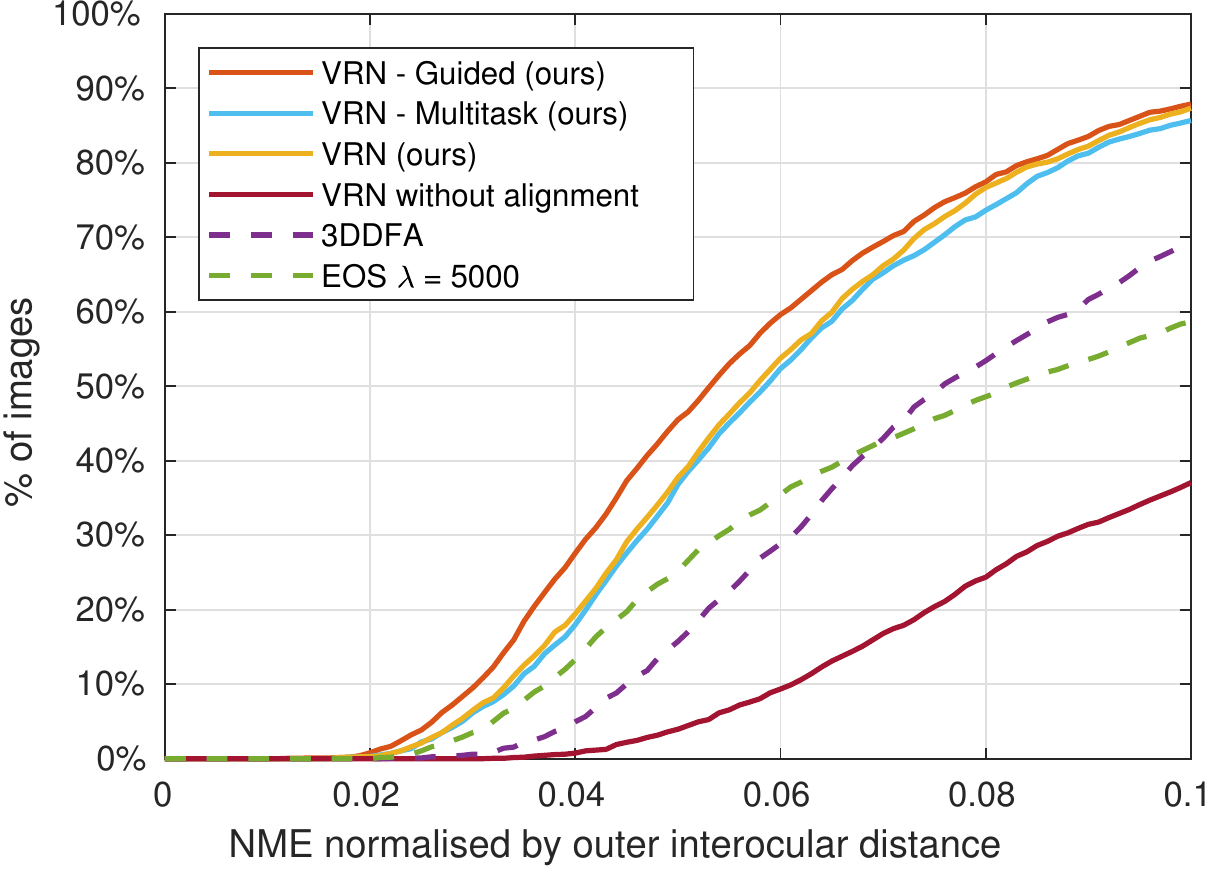}
  \label{roc:aflw2000}
\end{subfigure}
\hspace{13mm}
\begin{subfigure}[t]{0.43\textwidth}
\centering
  \includegraphics[width=\linewidth]{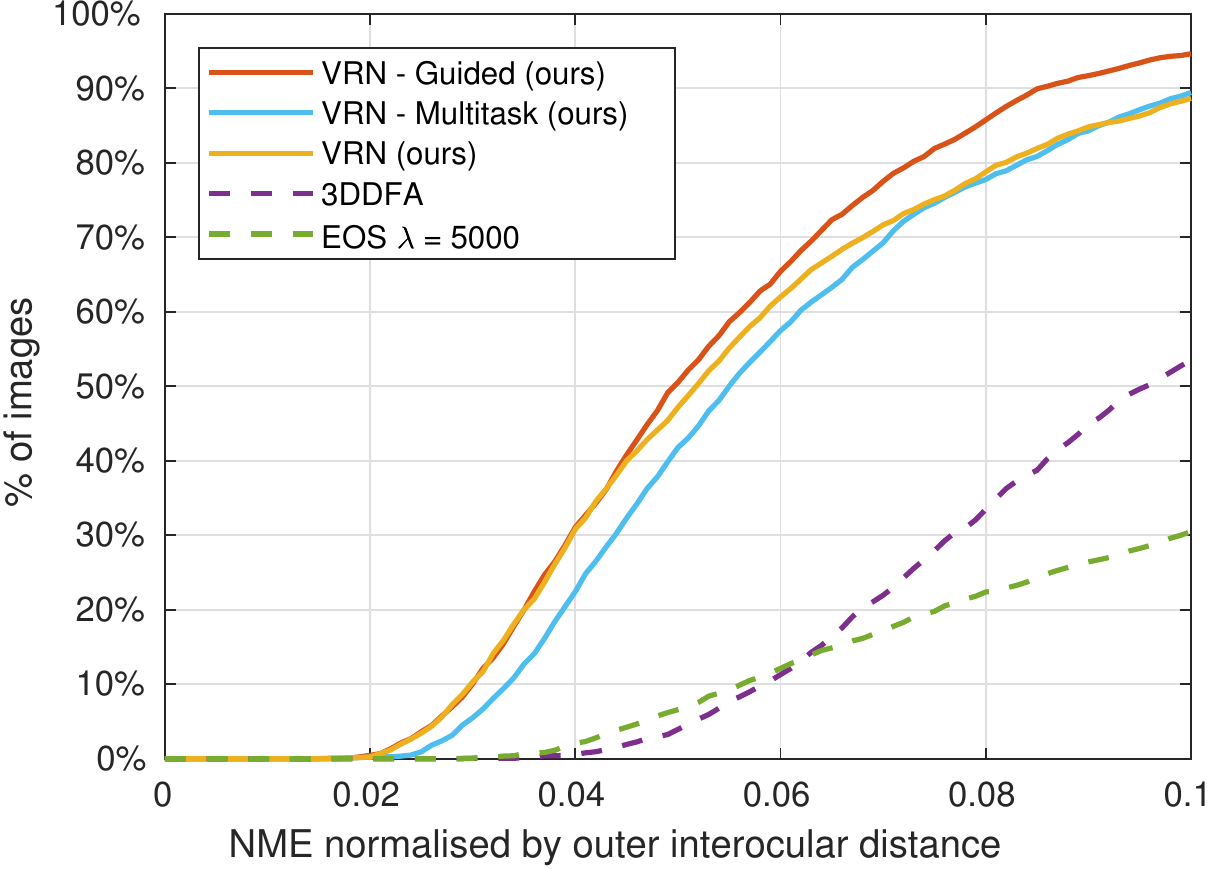}
  \label{roc:bu4dfe}
\end{subfigure}
\vspace{-13pt}
\caption{NME-based performance on in-the-wild ALFW2000-3D dataset (left) and renderings from BU-4DFE (right).  The proposed \textit{Volumetric Regression Networks}, and EOS and 3DDFA are compared.}
\label{roc:combined}
\vspace{-4mm}
\end{figure*}

\begin{figure}
\centering
  \includegraphics[width=0.92\linewidth]{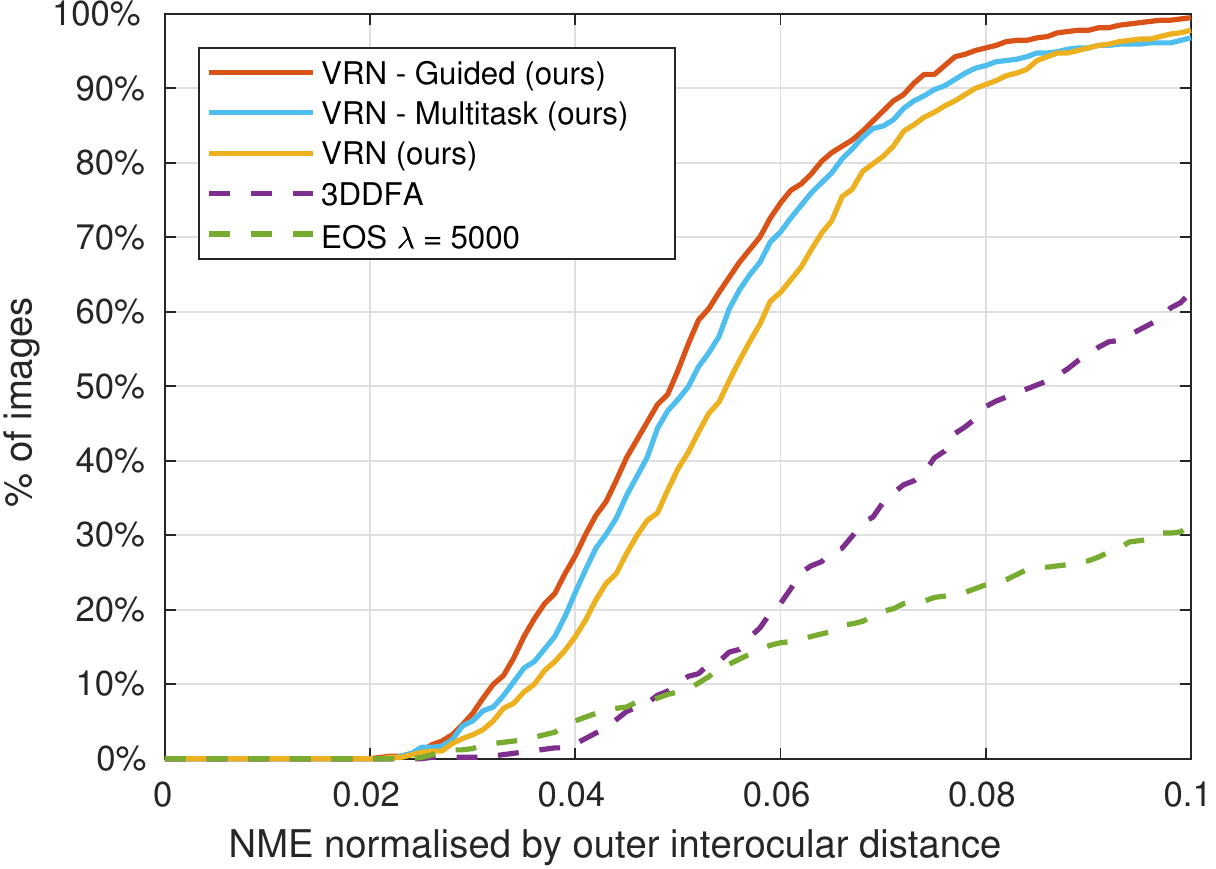}
\caption{NME-based performance on our large pose renderings of the Florence dataset. The proposed \textit{Volumetric Regression Networks}, and EOS and 3DDFA are compared.}
  \label{roc:florence}
\end{figure}

\begin{table}
  \caption{Reconstruction accuracy on AFLW2000-3D, BU-4DFE and Florence in terms of NME. Lower is better.}
  \label{tab:overview}
  \centering
  \small
\begin{tabular}{|l||c|c|c|}
  \hline
  \textbf{Method}   & \textbf{AFLW2000-3D} & \textbf{BU-4DFE} & \textbf{Florence} \\
  \hline\hline
  VRN & 0.0676   & 0.0600 & 0.0568   \\
  VRN - Multitask   & 0.0698        & 0.0625     & 0.0542        \\
  VRN - Guided    & \textbf{0.0637}   & \textbf{0.0555} & \textbf{0.0509}   \\
\hline
  3DDFA~\cite{zhu2016face}             & 0.1012   & 0.1227 & 0.0975   \\
  EOS~\cite{huber2016multiresolution}  & 0.0971   & 0.1560 & 0.1253   \\
  \hline
\end{tabular}
\vspace{-4mm}
\end{table}

We performed cross-database experiments only, on 3 different
databases, namely AFLW2000-3D, BU-4DFE, and Florence reporting the
performance of all the proposed networks (\textit{VRN}, \textit{VRN -
  Multitask} and \textit{VRN - Guided}) along with the performance of
two state-of-the-art methods, namely 3DDFA \cite{zhu2016face} and EOS
\cite{huber2016multiresolution}. Both methods perform 3DMM
fitting (3DDFA uses a CNN), a process completely bypassed by \textit{VRN}.

Our results can be found in Table~\ref{tab:overview} and
Figs.~\ref{roc:combined} and ~\ref{roc:florence}.
Visual results of the proposed \textit{VRN - Guided} on some very
challenging images from AFLW2000-3D can be seen in
Fig.~\ref{fig:aflw2000res}. Examples of failure cases along with a visual comparison between \textit{VRN} and \textit{VRN - Guided} can be found in the supplementary material. From these results, we can conclude the
following:
\begin{enumerate}
\item
Volumetric Regression Networks largely outperform 3DDFA and EOS on all datasets, verifying that directly regressing the 3D facial structure is a much easier problem for CNN learning.
\item
All VRNs perform well across the whole spectrum of facial poses, expressions and occlusions. Also, there are no significant performance discrepancies across different datasets (ALFW2000-3D seems to be slightly more difficult).
\item The best performing VRN is the one guided by detected landmarks
  (\textit{VRN - Guided}), however at the cost of higher computational
  complexity: \textit{VRN - Guided} uses another stacked hourglass network for
  landmark localization.
\item
\textit{VRN - Multitask} does not always perform particularly better than the plain VRN (in fact on BU-4DFE it performs worse), not justifying the increase of network complexity. It seems that it might be preferable to train a network to focus on the task in hand.
\end{enumerate}

\noindent{}Details about our experiments are as follows:

\textbf{Datasets.} \textbf{(a) AFLW2000-3D:} As our target
was to test our network on totally unconstrained images, we firstly
conducted experiments on the AFLW2000-3D~\cite{zhu2016face} dataset
which contains 3D facial meshes for the first 2000 images from AFLW
\cite{aflw2011}. \textbf{(b) BU-4DFE:} We also conducted
experiments on rendered images from BU-4DFE~\cite{yin2008high}. We
rendered each participant for both Happy and Surprised expressions
with three different pitch rotations between $-20$ and $20$
degrees. For each pitch, seven roll rotations from $-80$ to $80$
degrees were also rendered. Large variations in lighting direction and
colour were added randomly to make the images more challenging.   \textbf{(c) Florence:} Finally,
we conducted experiments on rendered images from the
Florence~\cite{masi2d3dFaceData} dataset. Facial images were rendered
in a similar fashion to the ones of BU-4DFE but for slightly different
parameters: Each face is rendered in 20 difference poses, using a
pitch of -15, 20 or 25 degrees and each of the five evenly spaced
rotations between -80 and 80. 

\textbf{Error metric}. To measure the accuracy of reconstruction for each face, we used the Normalised Mean Error (NME) defined as the average per vertex Euclidean distance between the estimated and ground truth reconstruction normalised by the outer 3D interocular distance:

\begin{equation}
\textrm{NME} = \frac{1}{N} \sum_{k=1}^{N} \frac{||\mathbf{x}_k-\mathbf{y}_{k} ||_{2} }{d}, \label{eq:err}
\end{equation}
where $N$ is the number of vertices per facial mesh, $d$ is the 3D
interocular distance and $\mathbf{x}_k$,$\mathbf{y}_k$ are vertices of the grouthtruth and predicted meshes. The error is calculated on the face region only
on approximately 19,000 vertices per facial mesh. Notice that when
there is no point correspondence between the ground truth and the
estimated mesh, ICP was used but only to establish the
correspondence, i.e. the rigid alignment was not used. If the rigid
alignment is used, we found that, for all methods, the error decreases
but it turns out that the relative difference in performance remains
the same. For completeness, we included these results in the
supplementary material.

\textbf{Comparison with
  state-of-the-art.} We compared against state-of-the-art 3D
reconstruction methods for which code is publicly available. These include the very recent methods of 3DDFA~\cite{zhu2016face},  and EOS~\cite{huber2016multiresolution}\footnote{For EOS we used
  a large regularisation parameter $\lambda = 5000$ which we found to
  offer the best performance for most images. The method uses 2D landmarks as input, so for the
  sake of a fair comparison a stacked hourglass for 2D landmark
  detection was trained for this purpose. Our tests were performed
  using v0.12 of EOS.}. 








\section{Importance of spatial alignment}
\label{sec:spatialimportance}

  The 3D reconstruction method described
  in~\cite{choy20163d} regresses a 3D volume of fixed orientation from one or more images
  using an LSTM. This is different to our approach of taking a single
  image and regressing a spatially aligned volume, which we believe is
  easier to learn. To explore what the repercussions of ignoring spatial alignment are, we trained a variant of \textit{VRN} which regresses a frontal version of the face, i.e. a face of fixed orientation as in ~\cite{choy20163d} \footnote{ We also attempted to train a network using the code
  from~\cite{choy20163d} on downsampled versions of our own
  volumes. Unfortunately, we were unable to get the network to learn anything.}.

\begin{figure}
\includegraphics[width=\linewidth]{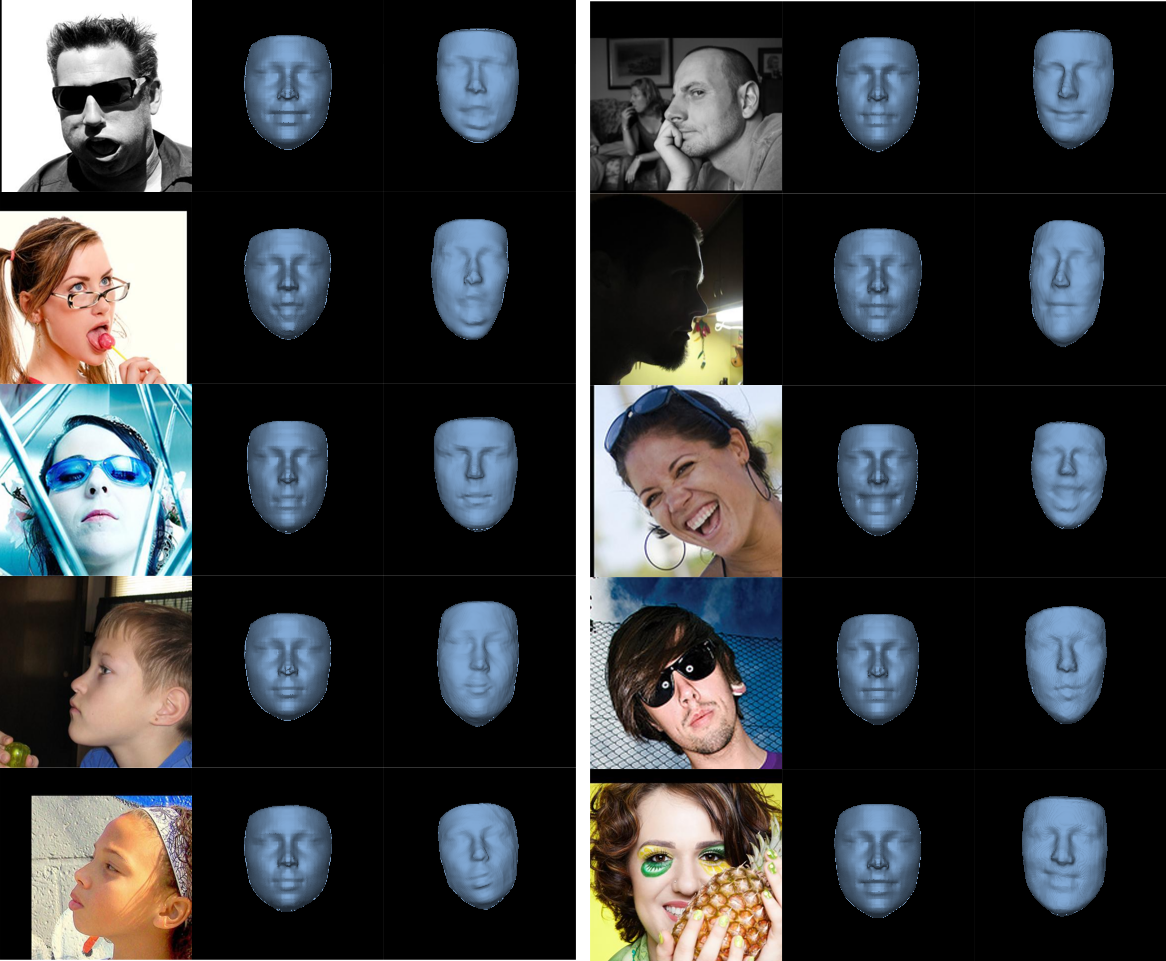}
\caption{Result from \textit{VRN without alignment} (second columns), and a frontalised output from \textit{VRN - Guided} (third columns).}
  \label{fig:frontal_visual}
\end{figure}

  Although this network produces a reasonable face, it can only capture diminished expression, and
  the shape for all faces appears to remain almost identical. This is very noticeable in Fig.~\ref{fig:frontal_visual}. Numeric comparison is shown in Fig.~\ref{roc:combined} (left), as \textit{VRN without alignment}. We believe that this further confirms that spatial alignment is of paramount importance when performing 3D reconstruction in this way.

\section{Ablation studies}

In this section, we report the results of experiments aiming to shed
further light into the performance of the proposed networks. For all
experiments reported, we used the best performing \textit{VRN -
  Guided}.

\begin{figure}
\centering
  \includegraphics[width=0.90\linewidth]{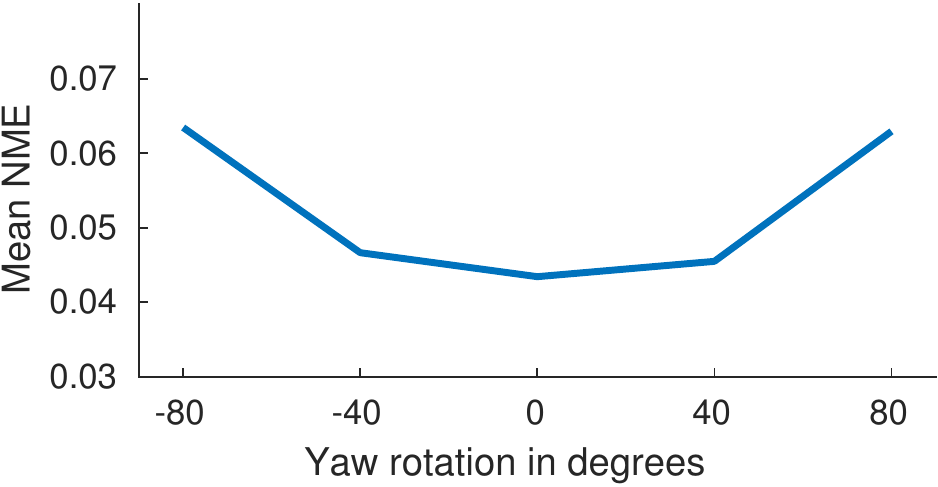}
  \caption{The effect of pose on reconstruction accuracy in terms of NME on the Florence dataset. The \textit{VRN - Guided} network was used.}
  \label{fig:effect_pose}
\end{figure}

\textbf{Effect of pose.} To measure the influence
of pose on the reconstruction error, we measured the NME for different
yaw angles using all of our Florence~\cite{masi2d3dFaceData}
renderings. As shown in Fig.~\ref{fig:effect_pose}, the performance of
our method decreases as the pose increases. This is to be expected,
due to less of the face being visible which makes evaluation for the
invisible part difficult. We believe that our error is still very low
considering these
poses. 

\begin{figure}
  \centering
  \includegraphics[width=0.9\linewidth]{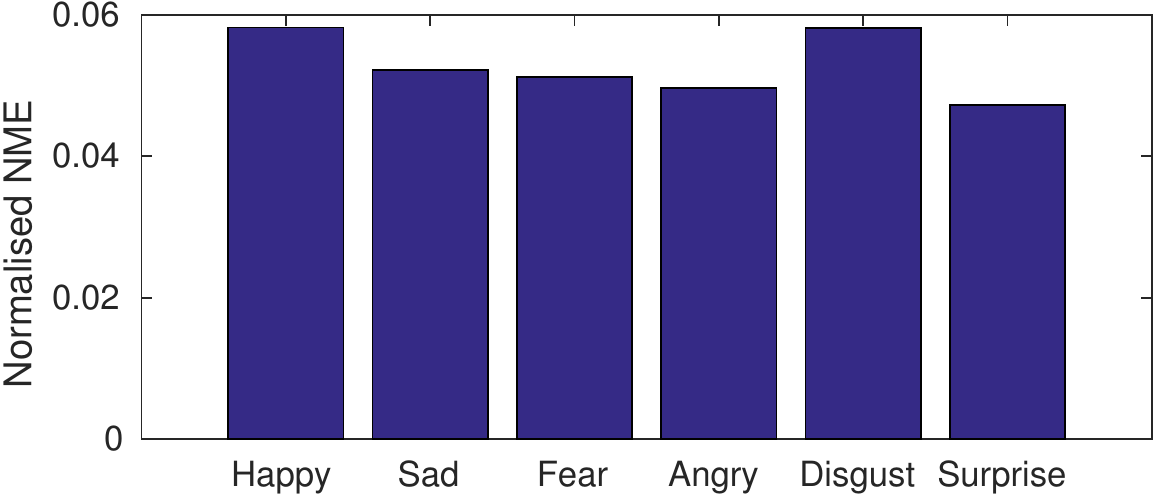}
  \caption{The effect of facial expression on reconstruction accuracy in terms of NME on the BU-4DFE dataset. The \textit{VRN - Guided} network was used.}
  \label{fig:effect_expression}
\end{figure}

\textbf{Effect of expression.} Certain expressions are usually considered harder to accurately reproduce in 3D face
reconstruction. To measure the effect of facial expressions on performance, we rendered frontal images in difference expressions from BU-4DFE (since Florence only
exhibits a neutral expression) and measured the performance for each expression. This kind of extreme acted facial
expressions generally do not occur in the training set, yet as shown in Fig.~\ref{fig:effect_expression}, the performance variation across different expressions is quite minor.

\textbf{Effect of Gaussian size for guidance.} We trained a
\textit{VRN - Guided}, however, this time,
the facial landmark detector network of the \textit{VRN - Guided} regresses larger Gaussians ($\sigma = 2$ as
opposed to the normal $\sigma = 1$). The performance of the 3D
reconstruction dropped by a negligible amount, suggesting that as long
as the Gaussians are of a sensible size, guidance will always help.

\section{Conclusions}

We proposed a direct approach to 3D facial reconstruction from a
single 2D image using volumetric CNN regression. To this end, we
proposed and exhaustively evaluated three different networks for
volumetric regression, reporting results that show that the proposed
networks perform well for the whole spectrum of facial pose, and can
deal with facial expressions as well as occlusions. We also compared
the performance of our networks against that of recent
state-of-the-art methods based on 3DMM fitting reporting large
performance improvement on three different datasets.
Future work may include improving detail and establishing a fixed correspondence from the isosurface of the mesh.

\section{Acknowledgements}

Aaron Jackson is funded by a PhD scholarship
from the University of Nottingham. We are grateful for access to the University of Nottingham
High Performance Computing Facility. Finally, we would like to express our thanks to Patrik
Huber for his help testing EOS~\cite{huber2016multiresolution}.





\clearpage
{\small
\bibliographystyle{ieee}
\bibliography{egbib}

\begin{thebibliography}{10}\itemsep=-1pt

\bibitem{masi2d3dFaceData}
A.~D. "Bagdanov, I.~Masi, and A.~Del~Bimbo.
\newblock The florence 2d/3d hybrid face datset.
\newblock In {\em Proc. of ACM Multimedia Int.’l Workshop on Multimedia
  access to 3D Human Objects (MA3HO’11)}. ACM, ACM Press, December 2011.

\bibitem{blanz1999morphable}
V.~Blanz and T.~Vetter.
\newblock A morphable model for the synthesis of 3d faces.
\newblock In {\em Computer graphics and interactive techniques}, 1999.

\bibitem{cao2014facewarehouse}
C.~Cao, Y.~Weng, S.~Zhou, Y.~Tong, and K.~Zhou.
\newblock Facewarehouse: A 3d facial expression database for visual computing.
\newblock {\em IEEE TVCG}, 20(3), 2014.

\bibitem{choy20163d}
C.~B. Choy, D.~Xu, J.~Gwak, K.~Chen, and S.~Savarese.
\newblock 3d-r2n2: A unified approach for single and multi-view 3d object
  reconstruction.
\newblock {\em arXiv preprint arXiv:1604.00449}, 2016.

\bibitem{eigen2015predicting}
D.~Eigen and R.~Fergus.
\newblock Predicting depth, surface normals and semantic labels with a common
  multi-scale convolutional architecture.
\newblock In {\em ICCV}, 2015.

\bibitem{eigen2014depth}
D.~Eigen, C.~Puhrsch, and R.~Fergus.
\newblock Depth map prediction from a single image using a multi-scale deep
  network.
\newblock In {\em NIPS}, 2014.

\bibitem{he2015deep}
K.~He, X.~Zhang, S.~Ren, and J.~Sun.
\newblock Deep residual learning for image recognition.
\newblock 2016.

\bibitem{huber2016multiresolution}
P.~Huber, G.~Hu, R.~Tena, P.~Mortazavian, W.~P. Koppen, W.~Christmas,
  M.~R{\"a}tsch, and J.~Kittler.
\newblock A multiresolution 3d morphable face model and fitting framework.

\bibitem{jourabloo2016large}
A.~Jourabloo and X.~Liu.
\newblock Large-pose face alignment via cnn-based dense 3d model fitting.
\newblock In {\em CVPR}, 2016.

\bibitem{kemelmacher20113d}
I.~Kemelmacher-Shlizerman and R.~Basri.
\newblock 3d face reconstruction from a single image using a single reference
  face shape.
\newblock {\em IEEE TPAMI}, 33(2):394--405, 2011.

\bibitem{kemelmacher2011face}
I.~Kemelmacher-Shlizerman and S.~M. Seitz.
\newblock Face reconstruction in the wild.
\newblock In {\em ICCV}, 2011.

\bibitem{aflw2011}
M.~Koestinger, P.~Wohlhart, P.~M. Roth, and H.~Bischof.
\newblock Annotated facial landmarks in the wild: A large-scale, real-world
  database for facial landmark localization.
\newblock In {\em First IEEE International Workshop on Benchmarking Facial
  Image Analysis Technologies}, 2011.

\bibitem{liu2016joint}
F.~Liu, D.~Zeng, Q.~Zhao, and X.~Liu.
\newblock Joint face alignment and 3d face reconstruction.
\newblock In {\em ECCV}, 2016.

\bibitem{long2015fully}
J.~Long, E.~Shelhamer, and T.~Darrell.
\newblock Fully convolutional networks for semantic segmentation.
\newblock In {\em CVPR}, 2015.

\bibitem{newell2016stacked}
A.~Newell, K.~Yang, and J.~Deng.
\newblock Stacked hourglass networks for human pose estimation.
\newblock In {\em ECCV}, 2016.

\bibitem{pavlakos2016coarse}
G.~Pavlakos, X.~Zhou, K.~G. Derpanis, and K.~Daniilidis.
\newblock Coarse-to-fine volumetric prediction for single-image 3d human pose.
\newblock {\em arXiv preprint arXiv:1611.07828}, 2016.

\bibitem{paysan20093d}
P.~Paysan, R.~Knothe, B.~Amberg, S.~Romdhani, and T.~Vetter.
\newblock A 3d face model for pose and illumination invariant face recognition.
\newblock In {\em AVSS}, 2009.

\bibitem{pfister2015flowing}
T.~Pfister, J.~Charles, and A.~Zisserman.
\newblock Flowing convnets for human pose estimation in videos.
\newblock In {\em ICCV}, 2015.

\bibitem{richardson2016learning}
E.~Richardson, M.~Sela, R.~Or-El, and R.~Kimmel.
\newblock Learning detailed face reconstruction from a single image.
\newblock {\em arXiv preprint arXiv:1611.05053}, 2016.

\bibitem{romdhani2005estimating}
S.~Romdhani and T.~Vetter.
\newblock Estimating 3d shape and texture using pixel intensity, edges,
  specular highlights, texture constraints and a prior.
\newblock In {\em CVPR}, 2005.

\bibitem{roth2016adaptive}
J.~Roth, Y.~Tong, and X.~Liu.
\newblock Adaptive 3d face reconstruction from unconstrained photo collections.
\newblock In {\em CVPR}, 2016.

\bibitem{sagonas2013semi}
C.~Sagonas, G.~Tzimiropoulos, S.~Zafeiriou, and M.~Pantic.
\newblock A semi-automatic methodology for facial landmark annotation.
\newblock In {\em CVPR-W}, 2013.

\bibitem{suwajanakorn2014total}
S.~Suwajanakorn, I.~Kemelmacher-Shlizerman, and S.~M. Seitz.
\newblock Total moving face reconstruction.
\newblock In {\em ECCV}, 2014.

\bibitem{tompson2015efficient}
J.~Tompson, R.~Goroshin, A.~Jain, Y.~LeCun, and C.~Bregler.
\newblock Efficient object localization using convolutional networks.
\newblock In {\em CVPR}, 2015.

\bibitem{tran2016regressing}
A.~T. Tran, T.~Hassner, I.~Masi, and G.~Medioni.
\newblock Regressing robust and discriminative 3d morphable models with a very
  deep neural network.
\newblock {\em arXiv preprint arXiv:1612.04904}, 2016.

\bibitem{tulsiani2016learning}
S.~Tulsiani, H.~Su, L.~J. Guibas, A.~A. Efros, and J.~Malik.
\newblock Learning shape abstractions by assembling volumetric primitives.
\newblock {\em arXiv preprint arXiv:1612.00404}, 2016.

\bibitem{yin2008high}
L.~Yin, X.~Chen, Y.~Sun, T.~Worm, and M.~Reale.
\newblock A high-resolution 3d dynamic facial expression database.
\newblock In {\em Automatic Face \& Gesture Recognition, 2008. FG'08. 8th IEEE
  International Conference on}, pages 1--6. IEEE, 2008.

\bibitem{zhu2016face}
X.~Zhu, Z.~Lei, X.~Liu, H.~Shi, and S.~Z. Li.
\newblock Face alignment across large poses: A 3d solution.
\newblock 2016.

\end{thebibliography}
}

\end{document}